\documentclass[conference]{IEEEtran}
\IEEEoverridecommandlockouts

\usepackage{cite}
\usepackage{amsmath,amssymb,amsfonts}
\usepackage{algorithmic}
\usepackage{graphicx}
\usepackage{textcomp}
\usepackage{xcolor}
\usepackage{multirow}
\usepackage{multicol}
\usepackage{float}
\usepackage{placeins}
\usepackage[table]{xcolor}
\definecolor{verylightgray}{gray}{0.92}

\usepackage{balance}
\usepackage{booktabs}
\usepackage{subcaption}
\usepackage{comment}
\usepackage{balance}
\def\method{\textsc{DyGEval}}

\newtheorem{definition}{Definition}[section]

\def\method{\textsc{DyGEval}}
\def\BibTeX{{\rm B\kern-.05em{\sc i\kern-.025em b}\kern-.08em
    T\kern-.1667em\lower.7ex\hbox{E}\kern-.125emX}}
\begin{document}

\title{Test-time GNN Model Evaluation on Dynamic Graphs}

\author{
\IEEEauthorblockN{Bo Li, Xin Zheng, Ming Jin, Can Wang, Shirui Pan\IEEEauthorrefmark{1}\thanks{\IEEEauthorrefmark{1}Corresponding author}}
\IEEEauthorblockA{School of Information and Communication Technology, Griffith University, Gold Coast, Australia}
\textit{\{bo.li, xin.zheng, ming.jin, can.wang, s.pan\}@griffith.edu.au}
}

\maketitle

\begin{abstract}
Dynamic graph neural networks (DGNNs) have emerged as a leading paradigm for learning from dynamic graphs, which are commonly used to model real-world systems and applications. However, due to the evolving nature of dynamic graph data distributions over time, well-trained DGNNs often face significant performance uncertainty when inferring on unseen and unlabeled test graphs in practical deployment. In this case, evaluating the performance of deployed DGNNs at test time is crucial to determine whether a well-trained DGNN is suited for inference on an unseen dynamic test graph. In this work, we introduce a new research problem: \textit{DGNN model evaluation}, which aims to assess the performance of a specific DGNN model trained on observed dynamic graphs by estimating its performance on unseen dynamic graphs during test time. Specifically, we propose a \textbf{\underline{Dy}}namic \textbf{\underline{G}}raph neural network \textbf{\underline{Eval}}uator, dubbed \textbf{\method}, to address this new problem. The proposed \method~involves a two-stage framework: (1) test-time dynamic graph simulation, which captures the training-test distributional differences as supervision signals and trains an evaluator; and (2) \method~development and training, which accurately estimates the performance of the well-trained DGNN model on the test-time dynamic graphs. Extensive experiments demonstrate that the proposed \method~serves as an effective evaluator for assessing various DGNN backbones across different dynamic graphs under distribution shifts.
\end{abstract}

\section{Introduction}

Dynamic graphs are widely used in real-world applications~\cite{li2025assemble, shen2025understanding, li2025oasis} such as social networks \cite{berger2006framework, greene2010tracking, sun2022aligning, song2019session}, financial transaction networks \cite{nascimento2021dynamic, zhang2021dyngraphtrans}, and traffic networks \cite{li2023dynamic, zhou2020foresee, zhou2020riskoracle, lan2022dstagnn}, where the graph topology and attributes evolve over time. Due to the complexity of both spatial and temporal correlation patterns, learning expressive dynamic graph representations poses substantial challenges~\cite{jin2023expressive}. 
This is particularly relevant in real-world web applications, where the web is continuously changing with the addition, removal, or update of pages and links. To accurately capture the ongoing evolution of the World Wide Web, dynamic graph learning needs to be adaptable and capable of operating in real-time, which is essential for supporting real-time web analytics and monitoring. To address this, dynamic graph neural networks (DGNNs) have emerged as powerful models, demonstrating outstanding performance in dynamic graph representation learning by integrating both spatial and temporal dependencies, effectively capturing the evolving nature of graphs over time.

\begin{figure*}
    \centering
    \includegraphics[width=0.95\textwidth]{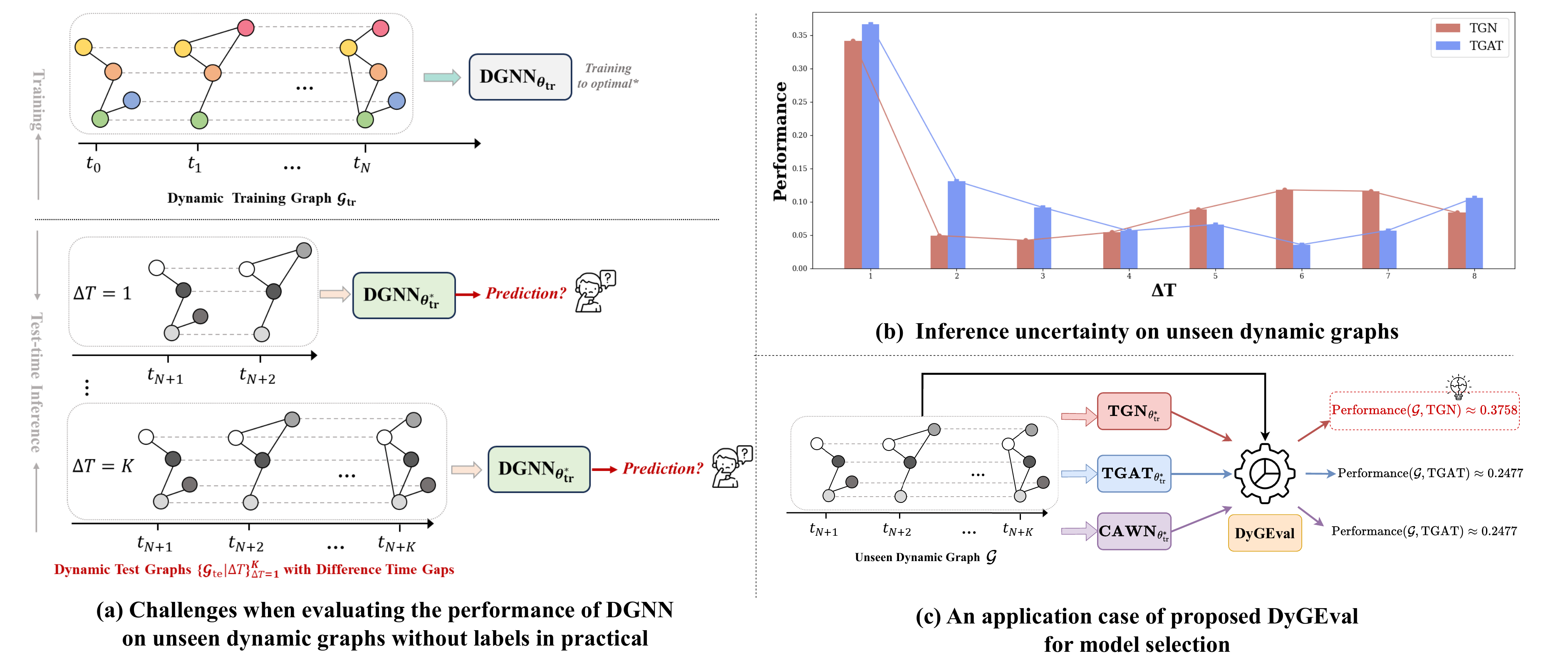}
    \caption{Illustration of \textit{necessity \& benefits} for test-time DGNN model evaluation: (a)
    The challenges faced when evaluating the performance of a well-trained DGNN on unseen test dynamic graphs without labels in practice; (b) Time gap varying caused test-time inference uncertainty; (c)An application case of the proposed test-time DGNN model evaluation problem for benefiting model selection.} 
    \label{fig:intro}
\end{figure*}

Despite the promising performance, DGNNs often face challenges when deployed in real-world scenarios, particularly in handling distribution shifts on time-evolving dynamic graphs~\cite{zhang2022dynamic}. As shown in Figure~\ref{fig:intro} (a), given a well-trained DGNN, without access to labels of the test dynamic graphs, it becomes challenging to assess the correctness of the DGNN's predictions, especially when distribution shifts arise due to temporal gaps.
For instance, in recommendation systems where dynamic graphs capture user-item interactions, changes in user features (e.g., an increase in income) can significantly alter user preferences and behaviors (e.g., a shift towards higher-priced items). Consequently, a DGNN model trained on earlier feature distributions (e.g., lower income) may struggle to deliver relevant recommendations at test time, as it relies on outdated interactions (such as purchases of lower-cost items)~\cite{yuan2024environment,wang2022causal}.

Hence, understanding and evaluating the performance of well-trained DGNN models during test time is crucial for measuring their effectiveness in real-world unseen dynamic graphs. Specifically, in Figure~\ref{fig:intro} (b), we test the performance of prevalent well-trained DGNNs, i.e., TGN~\cite{rossi2020temporal} and TGAT~\cite{xu2020inductive}, and CAWN~\cite{wang2021inductive},
on unseen dynamic graphs as the time gap $\Delta T$ gradually increases. Our fascinating findings reveal that in dynamic graphs, performance degradation due to distribution shifts is primarily linked to varying and progressively increasing time gaps between interactions. Larger time gaps increase the likelihood of shifts in both node features and graph topology, indicating larger distribution shifts between the training and test dynamic graphs.

To provide reliable guidance in test-time inference among different well-trained DGNNs for unseen dynamic graphs, in this work, we first propose a new research problem, i.e., test-time DGNN model evaluation, aimed at assessing the performance of well-trained DGNNs on unseen test dynamic graphs with time gap dominated distribution shift issues.
Given a well-trained DGNN model and an unseen test dynamic graph, the evaluation process directly outputs the performance of the well-trained DGNN model based on this specific test dynamic graph. 
This benefits model designers to confidently assess which well-trained DGNNs will perform better on future dynamic graphs, while also enabling users to evaluate the DGNNs at their disposal~\cite{jin2024timellm, jin2023large, huang2025themis, miao2025blindguard}. For example, as shown in Figure~\ref{fig:intro} (c), given a set of well-trained DGNNs, model evaluation provides designers and users with valuable insights, enabling them to make informed decisions when selecting the most suitable DGNN models for specific test dynamic graph inference.
To address this new research problem, we first identify two key challenges for developing effective test-time DGNN model evaluation methods. 
\textbf{Challenge 1:} Uncertain and varied distribution shifts, driven by evolving time gaps, make it challenging to accurately estimate the performance of well-trained DGNN models.
\textbf{Challenge 2:} Limited information, involving unseen and unlabeled test dynamic graphs and well-trained DGNNs, is available for developing a test-time DGNN model evaluator.

To address the above challenges, in this work, we propose a \textbf{\underline{Dy}}namic \textbf{\underline{G}}raph neural network \textbf{\underline{Eval}}uator, dubbed \textbf{\method}, for dealing with this new test-time DGNN model evaluation problem. The proposed \method~involves a two-stage framework: 
(1) \emph{Test-time dynamic graph simulation.} We build a set of dynamic graphs that maintain a distribution similar to the training set while incorporating diverse graph dynamics, characterized by time gap driven distribution shifts. This enables us to emulate potential unseen test dynamic graphs, effectively capturing and modeling complex and varied shifts between training and test graph distributions (\emph{Challenge 1}); 
(2) \emph{\method~development and training.} We design a graph transformer based DGNN evaluator, which first conducts discrepancy measurements to precisely calculate the training and simulated test dynamic graph distribution differences. Then, the proposed DGNN evaluator is trained to transmit the comprehensive discrepancies in node attributes, graph structures, temporal dynamics, and potential labels into accurate test-time DGNN performance estimations (\textit{Challenge 2}). Our contributions are listed as follows:

\(\bullet\) \textbf{New Problem}: We explore a new research problem: test-time DGNN model evaluation, which involves understanding and assessing the performance of historically well-trained DGNNs on future dynamic graphs in practical deployment scenarios, offering valuable insights for data-driven model selection at test time.

\(\bullet\) \textbf{Data-centric Solution}: We develop an effective solution in a novel two-stage \method~framework by comprehensively simulating and capturing discrepancies across diverse dynamic graph distributions, thereby enabling accurate in-service DGNN model evaluation at test time.

\(\bullet\) \textbf{Extensive Experiments}: We evaluate our method on typical DGNN models and real-world dynamic graphs, achieving minimal estimation gaps relative to the ground-truth DGNN performance demonstrating its outstanding effectiveness.

\section{Related Works}
\textbf{Graph Learning Model Evaluation.}
Our work is highly relevant to graph learning model evaluation under distribution shifts, facilitating a comprehensive understanding of graph learning model performance driven by varying graph data distribution shifts. Typically, Zheng et al. \cite{zheng2024gnnevaluator} proposed GNNEvaluator, first explored observing and predicting GNN model performance under various static graph data distribution shifts (e.g., domain changes) without relying on test-time ground truth labels. And other researches~\cite{zheng2024online, zheng2025test, liu2025test, zheng2024structure} went a further step by conducting static GNN model evaluation without requiring the training graph information.

\textbf{Temporal Graph Out-of-distribution (OOD) Generalization.} In addition, our work also relates to dynamic graph Out-of-distribution (OOD) Generalization, Out-of-distribution (OOD) generalization on temporal graphs aims to develop a TGNN model using several different but related source domains, enabling the model to generalize effectively to unseen target domains. Zhang et al. \cite{zhang2022dynamic} first investigate spatio-temporal distribution shifts in dynamic graphs, and their method effectively handles these shifts by discovering and fully utilizing invariant spatio-temporal patterns. Building on this work, Zhang et al. \cite{zhang2023out} apply the study of spatio-temporal distribution shifts in dynamic graphs to Human Albumin Prediction, providing accurate albumin predictions for Intensive Care Unit (ICU) patients during hospitalization. Furthermore, some studies have begun to concretize variant and invariant patterns. Zhang et al. \cite{zhang2024spectral} focus on spectral invariant learning for dynamic graphs under distribution shifts, effectively handling these shifts by capturing and utilizing both invariant and variant spectral patterns. Additionally, Yuan et al. \cite{yuan2024environment} explore environment-aware dynamic graph learning for OOD generalization, modeling complex coupled environments and exploiting spatio-temporal invariant patterns. The main difference between OOD generalization and our DGNN model evaluation lies in their objectives: OOD generalization focuses on generalizing OOD test samples at the data level, while our DGNN model evaluation focuses on assessing the performance of well-trained DGNNs on OOD data at the model level~\cite{liu2025graph}.
\section{Preliminary}

In this work, we focus on continuous-time temporal graphs, which we denote as timestamped edge streams and timestamps; i,e, in training set \(\mathcal{G}_{\text{tr}}= \{(e_i,t_i)\}_{i=1}^N\) and in test set \(\mathcal{G}_{\text{te}}= \{(e_{N+i},t_{N+i})\}_{i=1}^T\), where each interaction has two nodes at specific timestamp, e.g., \((e_i,t_i)=(\{(u_i,v_i\},t_i),t_i\in\mathbb{R}^+\). 
Moreover, in this task, we assume a covariate shift between \(\mathcal{G}_{\text{tr}}\) and \(\mathcal{G}_{\text{te}}\), where the distributional differences primarily occur in the node features and time. However, the output space of \(\mathcal{G}_{\text{te}}\) remains the same as that of \(\mathcal{G}_\text{tr}\).
Overall, we use \(N\) to denote the number of timestamps in the training dynamic graph, and we use \(T\) to denote the quantity of timestamps contains in the test dynamic graph.

\noindent\textbf{Dynamic Node Affinity Prediction with DGNNs}. 
Given two nodes $\{u_i, v_i\}$ at time \(t_i\) in a dynamic graph, the goal of dynamic node affinity prediction is to predict the interaction frequency vector \(\mathbf{y}_{i} \in \mathbb{R}^{C}\), where \(C\) denotes the label space, for node \(u_i\) with a set of candidate nodes $\mathcal{V}_{i}$ over a future period, i.e., \([t_i, t_{i+k}]\).
Each value in \(\mathbf{y}_{i}\) matches to a candidature \(v_i\in\mathcal{V}_{i}\).

\noindent\textit{\scalebox{0.6}{$\blacksquare$} Training.} Given a fully-observed dynamic training graph $\mathcal{G}_{\text{tr}}$ with corresponding ground-truth interaction labels $\mathbf{Y}_{\text{tr}}$, a DGNN model is trained for node affinity prediction task as:
\begin{equation}
    \min_{\boldsymbol{\theta}_{\text{tr}}^{*}} \mathcal{L}_{\text{ce}}(\hat{\mathbf{Y}}_{\text{tr}},\mathbf{Y}_{\text{tr}}), \, \text{where} \,\, \mathbf{Z}_{\text{tr}}, \hat{\mathbf{Y}}_{\text{tr}}=\text{DGNN}_{\boldsymbol{\theta}_{\text{tr}}^{*}}(\mathcal{G}_{\text{tr}}),
\end{equation}
where $\mathbf{Z}_{\text{tr}} \in \mathbb{R}^{N \times d}$ denotes the learned compact dynamic node embedding involving historical information, and $\hat{\mathbf{Y}}_{\text{tr}}$ denotes the predicted pseudo labels. By optimizing the node affinity prediction loss function \(\mathcal{L}_{\text{ce}}\), i.e., cross-entropy loss, we could get a well-trained DGNN with the optimal parameter \(\boldsymbol{\theta}_{\text{tr}}^{*}\).
Then, we utilize the normalized discounted cumulative gain (NDCG) metric \cite{jarvelin2002cumulated} to assess the performance of the dynamic node affinity prediction task, where NDCG indicates the measure ranking quality by examining the relative order of the top-ranked items, and a higher NDCG value indicates greater affinity. In this work, we use NDCG@10 to rank the top 10 destination nodes. The goal of dynamic node affinity prediction is to predict a node's property at any given timestamp \(t\). For example, to learn a function \(f: \mathbf{V}_t\rightarrow\mathcal{Y}\), \(\mathbf{V}_t\) is the set of nodes at time \(t\) and \(\mathcal{Y}\) is output space. Formally, given the observed evolution history of a temporal graph \(\mathcal{G}_t\) up to the current timestamp \(t\), the node affinity prediction task aims to predict the interaction frequency vector \(\mathbf{y}_t\) for a node \(u\) with a set of candidate nodes \(\mathbf{N}\) over a future period, i.e., \([t, t+k]\). Each value in \(\mathbf{y}_t\) matches to a candidate \(v\in\mathbf{N}\). The ground truth is:
\begin{equation}
    \mathbf{y}(u,v)=\frac{\sum_{t\leq{t_i}\leq{t+k}}w(u,v,t_i)}{\sum_{n\in\mathbf{N}}\sum_{t\leq{t_i}\leq{t+k}}w(u,n,t_i)}
\end{equation}
where \(w(u,v,t_i)\) is the weight of edge \((u,v)\) at timestamp \(t_i\).
We utilize the NDCG metric, which considers the relative order of elements. Before calculating NDCG, we first need to compute DCG (Discounted Cumulative Gain) and IDCG (Ideal Discounted Cumulative Gain). The DCG is calculated as follows:
\begin{equation}
    \text{DCG}=\sum_{i=1}^{k}\frac{2^{y_t[u,v_i]}-1}{\text{log}_2(i+1)}
\end{equation}
Similarly, the IDCG represents the ideal ranking of the DCG, where all relevant items are ranked in the best possible order. Hence, the NDCG is calculated as follow:
\begin{equation}
    \text{NDCG}=\frac{\text{DCG}}{\text{IDCG}}
\end{equation}

\noindent\textit{\scalebox{0.6}{$\blacksquare$} Test-time Inference.} Given an unknown and unlabeled test dynamic graph \(\mathcal{G}_{\text{te}}\), the well-trained DGNN conducts the standard test-time inference as: 
\begin{equation}
    \hat{\mathbf{Y}}_{\text{te}}= \text{DGNN}_{\boldsymbol{\theta}_{\text{tr}}^{*}}(\mathcal{G}_{\text{te}}).
\end{equation}
Therefore, we need a model evaluation technique to estimate how a well-trained DGNN performs on unseen dynamic graphs. Without ground truth values, it is impossible to calculate performance metrics like NDCG.
\begin{figure*}[!t]
    \centering
    \includegraphics[width=\textwidth]{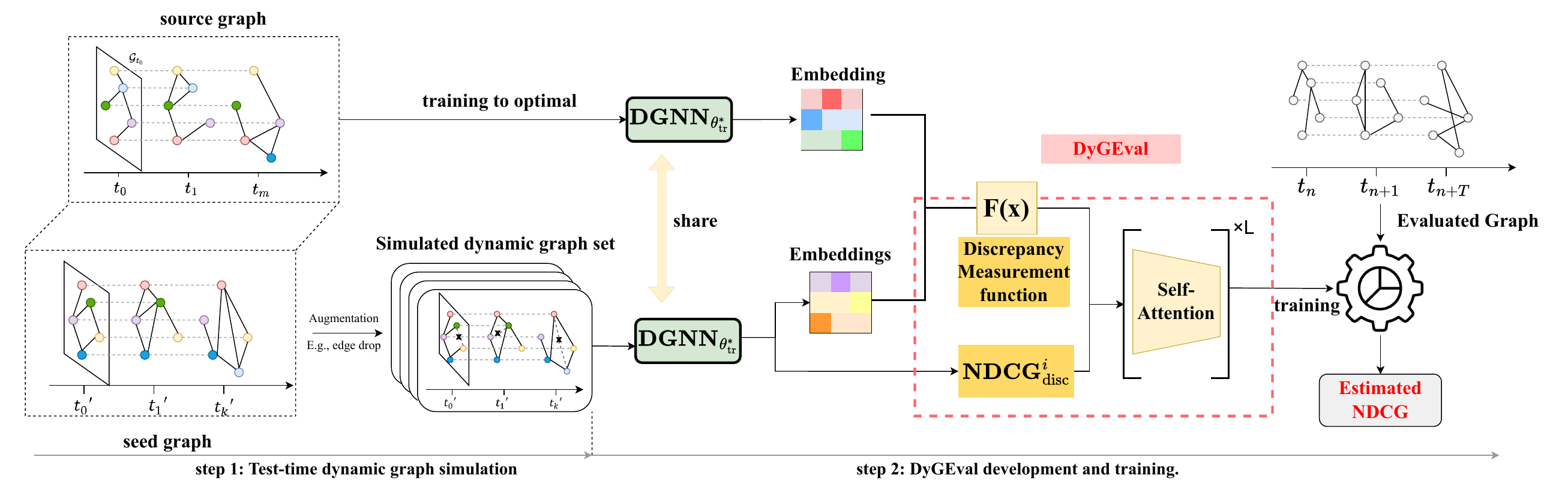}
    \caption{An overview of the two-stage pipeline for DGNN model evaluation consists of: (1) test-time dynamic graph simulation and (2) \method~development and training.}
    \label{fig:framework}
\end{figure*}

\section{Methodology}In this section, we first introduce the definition of the new research problem, followed by a brief introduction to our proposed \method~framework. Next, we provide a detailed explanation of how we conduct test-time DGNN model evaluation using our method.
\subsection{Problem Definition}
Given the scenarios where we cannot directly perform inference without labels and face time-gap-driven distribution shifts, we define a Test-time DGNN Model Evaluation task aimed at estimating the performance of a well-trained DGNN on unseen dynamic graphs.
\begin{definition}[\textbf{Test-time DGNN Model Evaluation}]
Given a fully-observed dynamic training graph \(\mathcal{G}_{\text{tr}}\) and its well-trained \(\text{DGNN}_{\boldsymbol{\theta}_{\text{tr}}^{*}}\). The unknown, unlabeled, and potentially distribution shifted test graph \(\mathcal{G}_{\text{te}}\) as inputs, test-time DGNN model evaluation aims to learn an approximate performance metric (Perf.Mec.) through an evaluator \(g_{\boldsymbol{\psi}} (\cdot)\) as follow:
\begin{equation}
    \text{Perf.Mec.} = g_{\boldsymbol{\psi}}(\text{DGNN}_{\boldsymbol{\theta}_{\text{tr}}^{*}}, \mathcal{G}_{\text{tr}}, \mathcal{G}_{\text{te}}),
\end{equation}
\end{definition}
where $\boldsymbol{\psi}$ is the learnable parameters.

In this work, we employ NDCG as the Perf.Mec. for evaluating well-trained DGNN models in the dynamic node affinity prediction task.  It is important to highlight that test-time DGNN model evaluation is a general research problem, encompassing various performance metrics applicable to a wide range of dynamic graph learning tasks.

\subsection{Framework of the Proposed \method}
\noindent As shown in Figure~\ref{fig:framework}, the proposed DGNN model evaluation framework consists of two stages: (1) Test-time dynamic graph simulation, and (2) \method~development and training.

\textbf{Stage 1}. Given the fully-observed dynamic training graph \(\mathcal{G}_{\text{tr}}\) with timestamps $\left\{t_1, t_2, \dots, t_N\right\}$, we firstly extract a dynamic seed graph \(\mathcal{G}_{\text{seed}}\) from it, containing timestamps $\left\{t_1', t_2' \dots, t_k'\right\}$. Afterward, we imply dynamic augmentations (e.g., Edgedrop) to the seed graph, so that we could obtain a simulated test dynamic graph set \(\mathcal{G}_{S}\). Then, the simulated test dynamic graph set \(\mathcal{G}_{S}\) and the observed dynamic training graph \(\mathcal{G}_{\text{tr}}\) are fed into the well-trained $\text{DGNN}_{\boldsymbol{\theta}_{\text{tr}^{*}}}$ to extract latent node embeddings with dynamic information i.e., \(\mathbf{Z}_s^{(i,*)}\) and \(\mathbf{Z}_{\text{tr}}^*\). At the same time, the well-trained $\text{DGNN}_{\boldsymbol{\theta}_{\text{tr}^{*}}}$ also provides node affinity prediction values based on the ground truth, resulting in new scalar NDCG values (\(\text{NDCG}_{\text{disc}}^i\)). 

\textbf{Stage 2}. In the second stage, we firstly calculate the distribution differences between training (\(\mathbf{Z}_{\text{tr}}^*\)) and simulated test dynamic graph (\(\mathbf{Z}_s^{(i,*)}\) ) by discrepancy measurement function. Then the DGNN evaluator is trained by the output calculated by discrepancy measurement function and new scalar NDCG values (\(\text{NDCG}_{\text{disc}}^i\)). Finally, the trained \method~could directly output the he node affinity prediction performance of test dynamic graph \(\mathcal{G}_{\text{te}}\).

\subsection{Test-Time Dynamic Graph Simulation}
\noindent The major challenge in developing DGNN model evaluation methods is the complexity and diversity of dynamic graph distribution discrepancies between potential real-world unseen test graphs and the observed training graphs. While practical in-service DGNNs perform well on known dynamic training graphs, inferring on diverse and varied unseen test graphs can lead to significantly uncertain performance for \method~.

In view of this, we propose constructing a diverse set of dynamic graphs to simulate the broad discrepancies in potential unseen graph distributions. Particularly, this set should have the following properties: (1) \textbf{Sufficient quantity}: it should contain a relatively large number of dynamic graphs with diverse node attributes and graph information; (2) \textbf{Represented discrepancy}: the node attributes of each graph should reflect its distributional shift from the dynamic training graph; (3) \textbf{Known prediction value}: each node should be labeled with its corresponding node affinity prediction value.

To satisfy property (1), we propose to simulate the synthesis of a wide variety of dynamic graphs, representing potential unseen test-time graph distributions that could be encountered in real-world scenarios. Concretely, we extract a dynamic seed graph \(\mathcal{G}_{\text{seed}}\) related to dynamic training graph \(\mathcal{G_{\text{tr}}}\). The key of \(\mathcal{G}_{\text{seed}}\) should exhibit as much distribution shift as possible from the dynamic training graph, while maintaining the same output space as \(\mathcal{G_{\text{tr}}}\) to satisfy the covariate shift consumption.

Next, \(\mathcal{G}_{\text{seed}}\) is augmented with a probability \(\mathbb{P}(p|\text{EdgeDrop})\), where \(p\in(0,1)\) is the augmentation ratio applied to \(\mathcal{G}_{\text{seed}}\). For example, \(p=0.3\) is dropping 30\% edges in \(\mathcal{G}_{\text{seed}}\). By doing this, we can get a set of simulated dynamic graphs which contain \(N\) simulated dynamic graphs represented as \(\mathcal{G}_{\mathrm{S}}=\{{g}_{\mathrm{s}}^i\}_{i=1}^N\). Each simulated dynamic graph is \(g_{\mathrm{s}}^i=\{(e_{t_i'},t_i')\}_{i=1}^K\) that contains \(K\) timestamps.

To satisfy the property (2), we fully leverage the outputs of node embeddings and node affinity predictions from well-trained \(\text{DGNN}_{\boldsymbol{\theta}_{\text{tr}}^{*}}\), incorporating different distribution shifts between training and test dynamic graphs into distinctive discrepancy representations. In particular, \(\mathbf{Z}_{\text{tr}}\) is the latent node embedding of training dynamic graph from \(\text{DGNN}_{\boldsymbol{\theta}_{\text{tr}}^{*}}\). \(\mathbf{Z}_{\mathrm{s}}^i\in\mathbb{R}^{M_i\times d}\) is the \(i\)-th latent node representation of simulated dynamic graph set generated by \(\text{DGNN}_{\boldsymbol{\theta}_{\text{tr}}^{*}}\). We define a discrepancy measurement function \(F(\cdot)\) to measure the distributional distance of the discrepancy simulated dynamic graph node attribute as follows:
\begin{equation} \label{similarityF}
    \mathbf{X}_{\text{disc}}^i = F(\mathbf{Z}_{\text{tr}}, \mathbf{Z}_{\mathrm{s}}^i) = \frac{\mathbf{Z}_{\mathrm{s}}^i \mathbf{Z}_{\text{tr}}^\mathrm{T}}{||\mathbf{Z}_{\mathrm{s}}^i||_2 \cdot ||\mathbf{Z}_{\text{tr}}||_2}.
\end{equation}
Here, \(\mathbf{X}_{\text{disc}}^i\in\mathbb{R}^{M_i\times N}\), where \(M_i\) denotes the number of nodes in the \(i\)-th simulated dynamic graph, represents the node-level distribution discrepancy with temporal information between each \(g_{\mathrm{s}}\) and \(\mathcal{G_{\text{tr}}}\). Each \(x_{u,v}^i\in \mathbf{X}_{\text{disc}}^i \) denotes the embedding discrepancy gap between node \(u\) in the \(i\)-th simulated dynamic graph and a node \(v\) in the dynamic training graph.

To satisfy property (3), known prediction values can be involved with the output of node property classification produced by \(\text{DGNN}_{\boldsymbol{\theta}_{\text{tr}}^{*}}\) on simulated dynamic graph \(\hat{\mathbf{Y}}_{\mathrm{s}}^i = \{\hat{y}_{\mathrm{s}}^j\}_{j=1}^{M_i}\). We calculate the node affinity prediction NDCG on the simulated dynamic graph given its ground truth node labels \(\mathbf{Y}_{\mathrm{s}}^i =\{{y_{\mathrm{s}}^j\}_{j=1}^{M_i}} \) as \(y_{\text{disc}}^i=\text{NDCG}(g_{\mathrm{s}}^i)\). Note that \(y_{\mathrm{disc}}^i\in\mathbb{R}\) is a continuous scalar which denotes node affinity prediction value under specific \(\text{DGNN}_{\boldsymbol{\theta}_{\text{tr}}^{*}}\) with the interval of \((0,1)\).

\subsection{\method~Development and Training}
 By incorporating all these properties with the discrepancy node attributes \(\mathbf{X}_{\text{disc}}^i\), the node affinity prediction label  \(y_{\text{disc}}^i\), we can derive the discrepancy dynamic graph set $\mathcal{D}=\{d_i\}_{i=1}^N$ between the test-time simulated dynamic graphs and the training graphs, where it can be denoted as follows:
\begin{equation} \label{discrepancy}
    \mathcal{D}=\{d_i\}_{i=1}^N, \textrm{where}\: d_{i} = (\mathbf{X}_{\text{disc}}^i, y_{\text{disc}}^i).
\end{equation}

Note that for dynamic graphs, it is challenging to directly and accurately determine the structure (dynamic interactions) between two dynamic graphs across continuous timestamps, that is why we do not explicitly incorporate dynamic graph structure discrepancy in the $\mathcal{D}=\{d_i\}_{i=1}^N$.

In light of this, for effectively training \textsc{\method}, we propose a graph transformer based architecture as the test-time dynamic graph evaluator, where every node representation from the output of our \method~can incorporate graph-level information without additional structure encoding \cite{ying2021transformers}.

Specifically, given our constructed discrepancy set \( \mathcal{D}=\{d_i\}_{i=1}^N\) and \(d_{i} = (\mathbf{X}_{\text{disc}}^i, y_{\text{disc}}^i)\) in Eq.~(\ref{discrepancy}), our proposed \method~can be constructed by Graphormer-based model:
\begin{equation}
    \mathbf{h}'^{(l)}=\mathbf{MHA}(\text{LN}(\mathbf{h}^{(l-1)}))+\mathbf{h}^{(l-1)},
\end{equation}
\begin{equation}
        \mathbf{h}^{(l)}=\mathbf{FFN}(\text{LN}(\mathbf{h}'^{(l)}))+\mathbf{h}'^{(l)}.
\end{equation}
Here \(\text{MHA}\) denotes multi-head attention, \(\text{FFN}\) represents the feed-forward blocks and \(\text{LN}\) stands for the layer normalization.

Next, by mapping the embeddings to scalar node affinity prediction values for the entire set, the objective function is:
\begin{equation}
    \min_{\boldsymbol{\psi}}\sum_{i=1}^N \mathcal{L}_{\text{MSE}}(g_{\boldsymbol{\psi}}(\mathbf{X}_{\text{disc}}^i),y_{\text{disc}}^i).
\end{equation}
Here, \(\mathcal{L}_{\text{MSE}}\) is the mean square error loss between objective function prediction of \(\mathbf{X}_{\text{disc}}^i\), and its label value \(y_{\text{disc}}^i\).
During the inference with the practical DGNN model, we have \textcircled{1} well-trained \(\text{DGNN}_{\boldsymbol{\theta}_{\text{tr}}^{*}}\) and \textcircled{2} unknown test dynamic graph \(\mathcal{G}_{\text{te}}=\{(e_{N+i},t_{N+i})\}_{i=1}^T\) without any information. The first is to compute the distance between the unseen dynamic test graph \(\mathcal{G}_{\text{te}}\) and training graph \(\mathcal{G_{\text{tr}}}\) based on Eq.~(\ref{similarityF}), yielding \(\mathbf{X}_d^{\text{te}}\). After this, the proposed \textsc{\method~} can produce the output of the node affinity prediction of \(\text{DGNN}_{\boldsymbol{\theta}_{\text{tr}}^{*}}\) on \(\mathcal{G}_{\text{te}}\) as:
\begin{equation}
    \text{NDCG}(\mathcal{G}_{\text{te}})=\hat{y}_d^{\text{te}}=g_{\boldsymbol{\psi}^*}(X_d^{\text{te}}).
\end{equation}
Here, \(\boldsymbol{\psi}^*\) represents the optimal \method~weight parameters trained by our discrepancy set \(\mathcal{D}\).
\begin{table*}[!t]
    \centering
    \caption{Mean absolute error (MAE $\downarrow$) performance comparison between different DGNN model evaluation baselines vs. our proposed \method~on three real-life datasets. The best results are bold and the second bests are underlined.}
    \resizebox{\linewidth}{!}{
    \begin{tabular}{l|cccc|cccc|cccc}
        \toprule
        \multirow{2}{*}{Methods}&\multicolumn{4}{c|}{tgbn-trade-TTE}&\multicolumn{4}{c|}{tgbn-genre-TTE}&\multicolumn{4}{c}{tgbn-reddit-TTE}\\  \cmidrule(r){2-5}  \cmidrule(r){6-9}  \cmidrule(r){10-13}&TGN&TGAT&CAWN&\textit{Avg.}&TGN&TGAT&CAWN&\textit{Avg.}&TGN&TGAT&CAWN&\textit{Avg.}\\
        \midrule
        Thres.\((\tau=0.5)\)& 0.1224& 0.1299& 0.2046& 0.1530 &0.0732& 0.1465& 0.1471& 0.1223&0.0726& 0.3649& 0.2964&0.2446\\
        Thres.\((\tau=0.7)\) & 0.1137& 0.3952& 0.1601& 0.2230&0.1061& 0.1468& 0.0460& 0.0996&0.0788& 0.2284&0.1266 &0.1446\\
        Thres.\((\tau=0.9)\) & 0.1006& 0.1519& 0.0717& 0.1081&0.0724& 0.1496& 0.0603& 0.0941&0.0512& 0.1588& 0.0723& 0.0941\\
        GNNEvaluator& \underline{0.0825}& \underline{0.0934}& \underline{0.0656}& \underline{0.0805}&\underline{0.0453}& \underline{0.1363}& \underline{0.0365}& \underline{0.0727}&\underline{0.0286}& \underline{0.1500}&\underline{0.0573} &\underline{0.0786}\\
        \rowcolor{verylightgray}
        \textbf{\method}& \textbf{0.0658}& \textbf{0.0821}& \textbf{0.0615}& \textbf{0.0698}&\textbf{0.0420}& \textbf{0.1007}& \textbf{0.0254}& \textbf{0.0560}&\textbf{0.0180}& \textbf{0.0727}&\textbf{0.0179} & \textbf{0.0362}\\
        \bottomrule
    \end{tabular}
    }
    \label{tab:MAE}
\end{table*}

\begin{table*}[!t]
    \centering
    \caption{Absolute error (AE $\downarrow$) of TGAT model evaluation methods on tgbn-trade-TTE.}
    \resizebox{0.9\textwidth}{!}{
    \begin{tabular}{l|ccccccccc}
        \toprule
          Methods&\(g_0\) &\(g_1\) &\(g_2\)&\(g_3\)&\(g_4\)&\(g_5\)&\(g_6\)&\(g_7\)&\textit{Avg.}\\
         \midrule
          Thres.\((\tau=0.5)\) & 0.0969&0.1637&0.0716&0.1114&0.1465&0.1368&0.1672&0.1458&0.1299\\
          Thres.\((\tau=0.7)\)&\underline{0.4031}&\underline{0.1425}&0.3778&0.4176&0.4527&0.4430&0.4734&0.4520&0.3952\\
          Thres.\((\tau=0.9)\)&0.1262&\textbf{0.1334}&	0.1009&0.1407&0.1758&0.1661&0.1965&0.1751&0.1519\\
          GNNEvaluator&0.0482&0.2124&	\underline{0.0229}&	\underline{0.0627}&\underline{0.0979}&	\underline{0.0881}&\underline{0.1177}&	\underline{0.0971}&	\underline{0.0934}\\
          \rowcolor{verylightgray}
          \textbf{\method~(TGAT)} &\textbf{0.0331}&	0.2275&	\textbf{0.0078}&	\textbf{0.0476}&	\textbf{0.0827}&	\textbf{0.0730}&	\textbf{0.1034}&\textbf{0.0820}&\textbf{0.0821}\\
          \bottomrule
    \end{tabular}
    }
    \label{tab:tgbn-trade-tgat}
\end{table*}

\begin{table*}[!t]
    \centering
    \caption{Absolute error  (AE $\downarrow$) of TGN model evaluation methods on tgbn-trade-TTE.}
    \resizebox{0.9\textwidth}{!}{
    \begin{tabular}{l|ccccccccc}
        \toprule
          Methods&\(g_0\) &\(g_1\) &\(g_2\)&\(g_3\)&\(g_4\)&\(g_5\)&\(g_6\)&\(g_7\)&\textit{Avg.}\\
         \hline
          Thres.\((\tau=0.5)\) & \textbf{0.0231}&0.1321&\underline{0.0388}&0.0463&0.0553&0.0914&0.0852&0.1130&0.0732\\
          Thres.\((\tau=0.7)\)&0.0710&0.1101&0.1215&0.0509&0.0659&0.1556&0.1402&0.1334&0.1061\\
          Thres.\((\tau=0.9)\)&0.03778&0.1083&\textbf{0.0364}&\textbf{0.0073}&\textbf{0.0049}&0.1341&0.1353&0.1175&0.0727\\
          GNNEvaluator&\underline{0.0240}&\underline{0.0459}&	0.0624&	0.0126&0.0194&	\underline{0.0653}&\underline{0.0666}&	\underline{0.0665}&	\underline{0.0453}\\
          \rowcolor{verylightgray}
          \textbf{\method~(TGN)}&0.0663&	\textbf{0.0337}&	0.1265&	\underline{0.0108}&	0.0278&	\textbf{0.0182}&\textbf{0.0308}&\textbf{0.0184}&\textbf{0.0420}\\
          \bottomrule
    \end{tabular}
    }
    \label{tab:tgbn-genre-TGN}
\end{table*}

\begin{table*}[!t]
    \centering
    \caption{Absolute error (AE $\downarrow$) of CAWN model evaluation methods on tgbn-trade-TTE.}
    \resizebox{0.9\textwidth}{!}{
    \begin{tabular}{l|ccccccccc}
        \toprule
          Methods&\(g_0\) &\(g_1\) &\(g_2\)&\(g_3\)&\(g_4\)&\(g_5\)&\(g_6\)&\(g_7\)&\textit{Avg.}\\
         \midrule
          Thres.\((\tau=0.5)\) & 0.1995&\textbf{0.0658}&0.1624&0.2108&0.2485&0.2362&0.2678&0.2463&0.2046\\
          Thres.\((\tau=0.7)\)&0.1399&\underline{0.1253}&0.1513&0.1891&0.1768&0.2083&0.1868&0.4520&0.1601\\
          Thres.\((\tau=0.9)\)&0.0578&0.3230&	0.0948&0.0464&0.0087&0.0210&\textbf{0.0105}&\underline{0.0110}&0.0717\\
          GNNEvaluator&\underline{0.0498}&0.3149&	\underline{0.0866}&	\underline{0.0382}&\underline{0.0005}&	\underline{0.0129}&\underline{0.0186}&	\textbf{0.0029}&	\underline{0.0656}\\
          \rowcolor{verylightgray}
          \textbf{\method~(CAWN)}&\textbf{0.0250}&	0.2902&	\textbf{0.0620}&	\textbf{0.0136}&	\textbf{0.0241}&	\textbf{0.0118}&0.0433&0.0218&\textbf{0.0615}\\
          \bottomrule
    \end{tabular}
    }
    \label{tab:tgbn-trade-CAWN}
\end{table*}

\begin{table}[!t]
    \centering
    \caption{Absolute error (AE $\downarrow$) of TGN model evaluation methods on eight test dynamic graphs of tgbn-trade-TTE. The best results are bold and the second bests are underlined.}
    \resizebox{\linewidth}{!}{
    \begin{tabular}{p{3.3cm}|>{\centering\arraybackslash}p{1.3cm}>{\centering\arraybackslash}p{1.3cm}>{\centering\arraybackslash}p{1.3cm}>{\centering\arraybackslash}p{1.3cm}}
        \toprule
         Methods&\(g_0\) &\(g_1\) &\(g_2\)\\
         \midrule
          Thres.\((\tau=0.5)\) & 0.9490&\textbf{0.0896}&0.1285\\
          Thres.\((\tau=0.7)\)&0.0870&0.2959&0.1213\\
          Thres.\((\tau=0.9)\)&0.0621&	\underline{0.1437}&	0.0965\\
          GNNEvaluator&\textbf{0.0485}&	0.3099&	\textbf{0.0162}\\
          \rowcolor{verylightgray}
          \textbf{\method~(TGN)}&\underline{0.0494}&	0.2311&	\underline{0.0544}\\
          \bottomrule
          &\(g_3\)&\(g_4\)&\(g_5\)\\
          \midrule
           Thres.\((\tau=0.5)\)&0.1357&0.1467&0.1195\\
          Thres.\((\tau=0.7)\)&0.1285&\underline{0.1162}&\underline{0.0367}\\
          Thres.\((\tau=0.9)\)&	0.1038&	0.1404&	0.1065\\
          GNNEvaluator&	\textbf{0.0081}&	\textbf{0.0233}&	0.0583\\
           \rowcolor{verylightgray}
           \textbf{\method~(TGN)}&	\underline{0.0208}&	0.1438&	\textbf{0.0257}\\
           \bottomrule
          &\(g_6\)&\(g_7\)&\textit{Avg.}\\
          \midrule
           Thres.\((\tau=0.5)\)&0.1320&0.1319&0.1224\\
          Thres.\((\tau=0.7)\)&\underline{0.0610}&\underline{0.0626}&0.1137\\
          Thres.\((\tau=0.9)\)&	0.0743&	0.0777&	0.1006\\
          GNNEvaluator&	0.1004&	0.0955&	\underline{0.0825}\\
           \rowcolor{verylightgray}
           \textbf{\method~(TGN)}&	\textbf{0.0013}&	\textbf{0.0001}&	\textbf{0.0658}\\
          \bottomrule
    \end{tabular}
    }
    \label{tab:tgbn-trade}
\end{table}
\section{Experiments}

 In all experiments, the DGNN models to be evaluated have been well-trained on observed dynamic training graphs and remain \textbf{fixed} during the entire test-time DGNN model evaluation process. We only vary the unseen test dynamic graph datasets and evaluate different types and backbones of the well-trained DGNNs. According to our proposed two-stage DGNN model evaluation framework, we do not have access to information about labels of the unseen dynamic test graphs. We only use their properties to obtain estimates for experimental illustration.
We investigate the following questions to verify the effectiveness of the proposed method.

\begin{itemize}
    \item \textbf{Q1:} How does the proposed \method~perform in predicting node properties for well-trained DGNNs?
    \item \textbf{Q2:} How does proposed \method~perform when conducting an ablation study for \method~component?
    \item \textbf{Q3:} How does proposed \method~perform when using different discrepancy measurement functions?
    \item \textbf{Q4:} How does proposed \method~perform when using various architecture?
\end{itemize}

\subsection{Experiment Settings}

\subsubsection{Datasets}
Considering we are the first test-time DGNN model evaluation method on the dynamic graphs, existing datasets can not be directly used for this new research problem, especially under the dynamic graph distribution shift scenarios. Therefore, we first build a new test protocol for our proposed research problem. Specifically, we start from three real-world graph datasets: tgbn-trade, tgbn-genre, and tgbn-reddit, all of which are part of the temporal graph benchmark dataset \cite{huang2023temporal}. Each dataset contains a different number of timestamps. We keep the original training set of these benchmarks for training the DGNN models.
For testing, we modify the original test graphs by introducing different starting timestamps, creating new test-time distribution-shifted dynamic graphs. These are used to evaluate our proposed \method.
For each dataset, we create eight test graphs, i.e., $\left\{g_0, g_1, \cdots, g_7\right\}$, leading to the variant of this dataset, named tgbn-trade-TTE, tgbn-genre-TTE, and tgbn-Reddit-TTE, where TTE denoting test-time evaluation.

In this task we evaluate our proposed \method~using the following scenario: \((t_0, t_1, \dots, t_m) \rightarrow (t_n, t_{n+1},\dots,t_{n+T})\). The left side represents the dynamic training graph, while the right side indicates the prediction made by \method~on the subsequent unseen dynamic test graph from the future. However, the performance of the test dynamic graph is fixed, and variations in the time length of the test graph do not significantly impact its performance. Therefore, we observe that when we introduce different beginning timestamps, the performance of the test dynamic graph on DGNN models trained with dynamic training graphs can change significantly. In this manner, we select seven different beginning timestamps to vary the performance of the test dynamic graphs on well-trained DGNNs. The original test dynamic graph, without any added timestamps, is represented by \(g_0\). Through this approach, we obtain \(\{g_0,g_1,\cdots,g_7\}\).
It is important to note that throughout all stages of \method~construction and inference, we do not access any label information from the unseen dynamic graph. The details of dataset statistics are in Table \ref{tab:dataset}.
\begin{table}[h]
    \centering
    \caption{Statistical details of the experimental datasets.}
    \resizebox{\linewidth}{!}{
    \begin{tabular}{cccccc}
        \toprule
         Scale&Name&\#Nodes&\#Edges&\#Steps&Metric\\
         \midrule
         \textcolor{red}{small}& \textcolor{blue}{tgbn-trade}&255&468,245&32&NDCG@10\\
         \textcolor{green}{medium}&\textcolor{blue}{tgbn-genre} & 1,505& 17,858,395&133,758& NDCG@10\\
         \textcolor{yellow}{large} &\textcolor{blue}{tgbn-reddit} & 11,766 & 27,174,118 & 21,889,537 &NDCG@10 \\
        \bottomrule
    \end{tabular}}
    \label{tab:dataset}
\end{table}.

\subsubsection{DGNN Models and Evaluation}
We evaluate three commonly used DGNN models: the memory-based model TGN \cite{rossi2020temporal}, the attention-based model TGAT \cite{xu2020inductive}, and the causal walk-based model CAWN \cite{wang2021inductive}. Each model is trained on the dynamic training set of the observed dynamic graph until it achieves the best node affinity prediction on its validation set, following the standard training process. We report the Mean Absolute Error (MAE), which represents the average absolute difference between the ground truth values on the unseen dynamic graph and the estimated node affinity prediction values from our model and the baselines on the same graph. A smaller MAE indicates better DGNN model evaluation performance.
\subsubsection{Baseline Methods}
Since our work is the first to address test-time evaluation for DGNN models, there is no directly comparable baseline. To ensure a fair evaluation, we consider two adapted methods. First, we extend the recently proposed static GNN evaluation method, \underline{GNNEvaluator}~\cite{zheng2024gnnevaluator}, to the dynamic graph setting. Unlike static GNN evaluation, dynamic graphs introduce an additional temporal dimension, making edge selection for evaluation non-trivial. To address this, we adapt GNNEvaluator by incorporating edges from the last timestamp, thereby transforming dynamic graphs into a static format. Specifically, GNNEvaluator constructs a discrepancy graph set $\mathcal{G}_{\text{disc}}=\{g_{\text{disc}}^i\}_{i=1}^K$ through meta-graph augmentation, where each discrepancy graph is defined by
\begin{equation}
    \mathbf{X}_{\text{disc}}^i = D(\mathbf{Z}_{\text{meta}}^{(i,*)},\mathbf{Z}_S^*)=\frac{\mathbf{Z}_{\text{meta}}^{(i,*)}{\mathbf{Z}_S^*}^\text{T}}{||\mathbf{Z}_{\text{meta}}^{(i,*)}||_2\cdot||\mathbf{Z}_S^*||_2}
\end{equation}
\begin{equation}
        y_{\text{disc}}^i=\text{Acc}(g_{\text{meta}}^i)=\frac{\sum_{j=1}^{M_i}(\hat{y}_{\text{meta}(i,*)}^j==y_{\text{meta}(i)}^j)}{M_i}
\end{equation}
Here, the adjacency matrix of each discrepancy graph $\mathbf{A}_{\text{disc}}^i$ is derived from the last-timestamp edges of the dynamic graph.

Second, we compare our method against a \underline{threshold-based method}, which assumes that predictions with higher softmax scores are more reliable. This method estimates model accuracy by computing the affinity:
\begin{equation}
    \text{affinity} = \frac{\sum_{i=1}^{N}\mathbf{1}\{\text{max}(f_\theta(x_j'))>\tau\}}{N},
\end{equation}
where $\tau$ is a predefined threshold. For dynamic graphs, we set $\tau=\{0.5,0.7,0.9\}$ and calculate affinity values based on DGNN softmax logits. When affinity between two nodes exceeds a threshold, it indicates frequent interactions between them.

\subsection{DGNNs Model Evaluation Results}
The evaluation of DGNNs on the tgbn-trade-TTE, tgbn-genre-TTE, and tgbn-reddit-TTE datasets reveals that the performance of our proposed evaluation method, \method, consistently outperforms baseline methods across different DGNN models and varying dynamic test graph configurations.

When looking at Table \ref{tab:MAE}, which presents the Mean Absolute Error (MAE) for different DGNN model evaluation baselines across three datasets, \method~consistently delivers the lowest MAE scores. Specifically, Table \ref{tab:tgbn-trade-tgat}, \ref{tab:tgbn-trade-CAWN}, \ref{tab:tgbn-genre-TGN} present the absolute error across different datasets and starting timestamps. Similar trends are observed across the tgbn-genre-TTE and tgbn-reddit-TTE datasets, where \method~continues to outperform the baselines. A deeper analysis of the results for individual models shows that \method~exhibits robustness across different dynamic models and datasets. For example, in the tgbn-genre-TTE dataset, \method~achieves an MAE of 0.0560 on average, compared to 0.0727 for GNNEvaluator. This trend is consistent across different DGNN models, indicating that \method~is effective regardless of the underlying DGNN architecture. On the tgbn-reddit-TTE dataset, we observe similar behavior, with \method~recording an average MAE of 0.0362, outperforming other baselines, including GNNEvaluator and threshold-based methods. \method's ability to maintain low errors across various time steps.

To more clearly illustrate the absolute gap between the estimated NDCG and GT, we also present more fine-grained results, i.e., absolute error (AE), related to Table \ref{tab:tgbn-trade}, we observe the absolute error results for various baselines on the tgbn-trade-TTE dataset. The results demonstrate that \method~achieves the lowest absolute error across most of the dynamic test graph configurations. For instance, if we only evaluate on test dynamic graph \(g_0\) , \method~yields an error of 0.0494, significantly lower than the other baseline methods such as Threshold \((\tau=0.5)\), which shows the highest error of 0.9490. Moreover, \method~outperforms GNNEvaluator in multiple configurations, as seen with a lower average error of 0.0658 compared to 0.0825 for GNNEvaluator.

Additionally, Table \ref{tab:rank} shows the relevance and NDCG of the \method~framework in evaluating various DGNN models by comparing their ranks across different datasets. The key insight here is when a model (such as TGN) and its evaluation method (e.g., \method~(TGN)) share the same rank, it indicates \method~successfully captures the model's true performance and provides an accurate evaluation. For example, In tgbn-trade-TTE dataset, TGN holds a rank of 3, and \method~(TGN) also ranks 2, aligning with TGN’s original performance. This shows that \method~accurately reflects the performance of the model. Similarly, in tgbn-genre-TTE dataset, TGN and \method~(TGN) both hold the second rank, reinforcing that \method~provides an accurate and consistent evaluation for TGN. For TGAT, the evaluation method \method~(TGAT) achieves the same rank of one across all three datasets, demonstrating that \method~accurately represents TGAT's superior performance.

These consistent rankings across models and their \method~evaluations suggest \method~can accurately capture the ranking and relevance of a model's performance. When the ranks align, as seen with TGN and \method~(TGN), or TGAT and \method~(TGAT), it illustrates that \method~delivers a reliable model evaluation.
\begin{table}[!t]
    \centering
    \caption{Comparative ranking of DGNN models and \method~framework across datasets.}
    \resizebox{\linewidth}{!}{
    \begin{tabular}{p{1.6cm}|>{\centering\arraybackslash}p{1.3cm}>{\centering\arraybackslash}p{1.3cm}|>{\centering\arraybackslash}p{1.3cm}>{\centering\arraybackslash}p{1.3cm}}
    \toprule
    \makebox[0.2\linewidth][l]{Dataset} & \multicolumn{4}{c}{tgbn-trade-TTE}\\
    \midrule
    \multirow{2}{*}{Metrics}&\multicolumn{2}{c|}{GT}&\multicolumn{2}{c}{\cellcolor{verylightgray}\textbf{\method~(Ours)}}\\\cmidrule(r){2-3}\cmidrule(r){4-5}
    &NDCG&\textit{rank}&\cellcolor{verylightgray}NDCG&\cellcolor{verylightgray}\textit{rank}\\ 
    \midrule
    TGN&0.1117&3 & \cellcolor{verylightgray}0.1197& \cellcolor{verylightgray}2\\
    TGAT&0.1137 &2& \cellcolor{verylightgray}0.1390& \cellcolor{verylightgray}1\\
    CAWN&0.1149 &1&\cellcolor{verylightgray} 0.0787&\cellcolor{verylightgray} 3\\
    \toprule
    Dataset & \multicolumn{4}{c}{tgbn-genre-TTE}\\
    \midrule
    \multirow{2}{*}{Metrics}&\multicolumn{2}{c|}{GT}&\multicolumn{2}{c}{\cellcolor{verylightgray}\textbf{\method~(Ours)}}\\\cmidrule(r){2-3}\cmidrule(r){4-5}
    &NDCG&\textit{rank}&\cellcolor{verylightgray}NDCG&\cellcolor{verylightgray}\textit{rank}\\
    \midrule
    TGN& 0.1229& 2& \cellcolor{verylightgray}0.1509& \cellcolor{verylightgray}2\\
    TGAT& 0.2355& 1 & \cellcolor{verylightgray}0.2062& \cellcolor{verylightgray}1\\
    CAWN& 0.0300& 3 &\cellcolor{verylightgray}0.0524& \cellcolor{verylightgray}3\\
    \toprule
    Dataset & \multicolumn{4}{c}{tgbn-reddit-TTE}\\
    \midrule
    \multirow{2}{*}{Metrics}&\multicolumn{2}{c|}{GT}&\multicolumn{2}{c}{\cellcolor{verylightgray}\textbf{\method~(Ours)}}\\\cmidrule(r){2-3}\cmidrule(r){4-5}
    &NDCG&\textit{rank}&\cellcolor{verylightgray}NDCG&\cellcolor{verylightgray}\textit{rank}\\
    \midrule
    TGN & 0.0873& 2&\cellcolor{verylightgray}0.0933&\cellcolor{verylightgray} 2\\
    TGAT&0.1316& 1&\cellcolor{verylightgray}0.1794& \cellcolor{verylightgray}1\\
    CAWN&0.0472& 3& \cellcolor{verylightgray}0.0365& \cellcolor{verylightgray}3\\
    \bottomrule
    \end{tabular}
    }
    \label{tab:rank}
\end{table}
\begin{table}[h]
    \centering
    \caption{Running time (in seconds) comparison between GNNEvaluator and our proposed \method~per epoch.}
    \resizebox{\linewidth}{!}{
    \begin{tabular}{p{2.5cm}|cc}
    \toprule
         Methods&tgbn-trade-TTE&tgbn-genre-TTE\\
         \midrule
         GNNEvaluator&0.0255&2.604 \\
         \rowcolor{verylightgray}
         \textbf{\method}&0.0897&3.882\\
         \bottomrule
    \end{tabular}
    }
    \label{tab:time}
\end{table}
\begin{figure*}[!t]
    \centering
    \includegraphics[width=\textwidth]{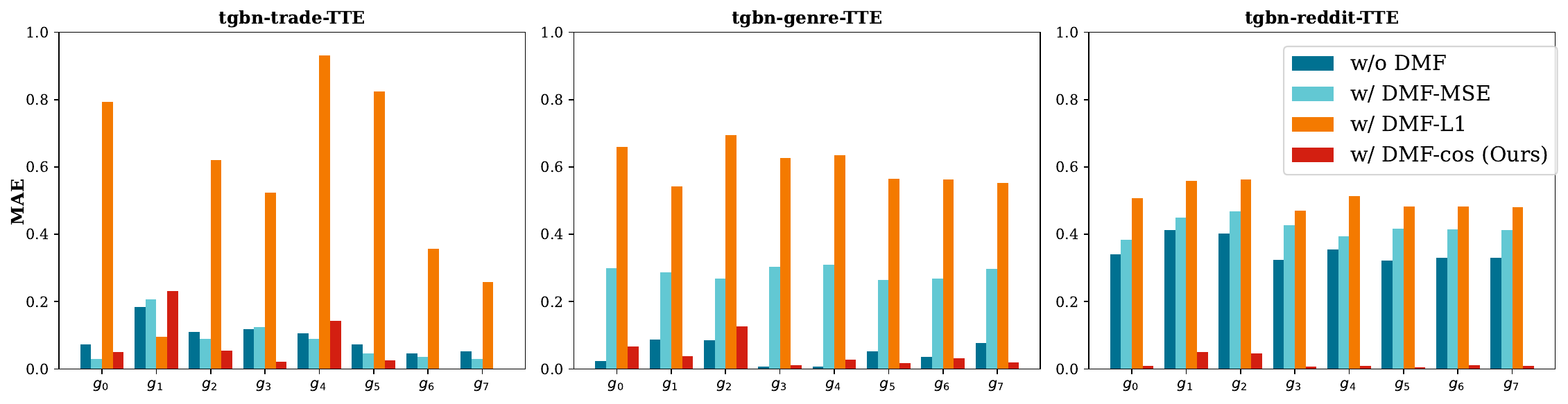}
    \caption{MAE of discrepancy measurement in \method~for TGN evaluation. Lower value ($\downarrow$) indicates better performance ($\uparrow$).}
    \label{fig:discrepancy measurement function}
\end{figure*}
\begin{figure*}[!t]
    \centering
    \includegraphics[width=\textwidth]{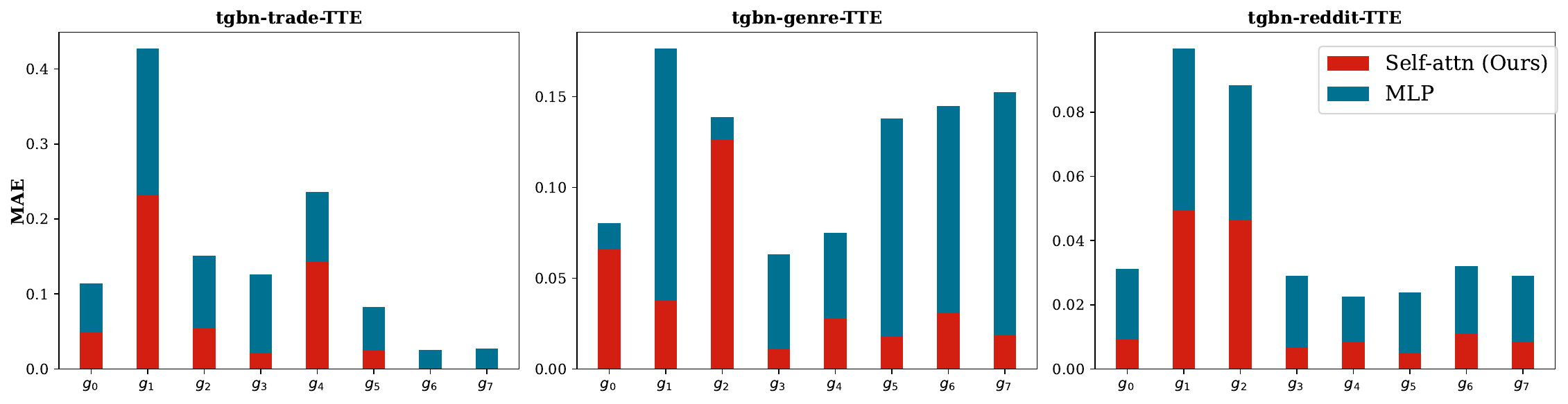}
    \caption{MAE of \method~backbones: Self-Attention vs. MLP for dynamic graph model evaluation. Lower value ($\downarrow$) indicates better performance ($\uparrow$).}
    \label{fig:backbone}
\end{figure*}
\subsection{Ablation Study}
We conduct an ablation study to assess the effectiveness of two submodules in the stage 2 of our proposed \method: (1) the discrepancy measurement function (DMF) and its variants; and (2) graph transformer based architecture.

\noindent \textbf{Effectiveness of Discrepancy Measurement}. We conduct an ablation study to assess the impact of the discrepancy measurement function in our proposed \method~pipeline. We remove the discrepancy measurement function and evaluate the model's performance using only the simulated dynamic graph to test the evaluation accuracy of the DGNNs on dynamic graphs. The figure presents the MAE of NDCG scores for node affinity prediction across three datasets: tgbn-trade-TTE, tgbn-genre-TTE, and tgbn-reddit-TTE, comparing \method~with the discrepancy measurement function and \method~without the discrepancy measurement function, as shown in Figure~\ref{fig:discrepancy measurement function}.

Overall, the results show that \method~trained with the discrepancy measurement function significantly outperforms the version without it, especially in tgbn-trade-TTE and tgbn-reddit-TTE. In tgbn-trade-TTE, the \method~(with DMF) value is closer to the GT, where the model achieves a higher NDCG score compared to the version without the discrepancy measurement function. In the tgbn-reddit-TTE dataset, the model without the discrepancy measurement function performs slightly better and more consistently, but still lags behind the version with the discrepancy measurement function.

The improved performance with the discrepancy measurement function is attributed to its ability to more effectively explore the latent node embedding space and capture the discrepancy between the training and testing dynamic graphs. Without the discrepancy measurement function, \method~relies solely on the simulated dynamic graph, which limits its ability to fully capture the hidden relationships between nodes and graph structures over time. 

Additionally, we analyze the effect of different discrepancy measurement functions—cosine similarity, L1 distance, and mean square error (MSE)—on the performance of the \method~for evaluating the TGN model. The model is trained on known dynamic training graphs and tested on future dynamic test graphs, with each discrepancy measurement function used to compute the discrepancy for training \method.

Overall, the analysis reveals that cosine similarity provides the best overall performance, as its NDCG values are more consistently aligned with the GT across multiple datasets. This suggests that cosine similarity is particularly effective in evaluating DGNNs compared to MSE and L1.

\noindent \textbf{Effectiveness of \method~Backbone.}
We explore the effect of different backbone architectures for \method~by replacing the self-attention mechanism, which is core to the Graphormer-like structure in \method~, with a simple MLP. The goal is to assess the influence of these two architectures—self-attention and MLP—on dynamic graph model evaluation across different datasets. The figure compares the MAE between output of \method~with self-attention (\method~(w Self-att)) and \method~with an MLP backbone (\method~(w MLP)).

Figure~\ref{fig:backbone} demonstrates the MAE of \method~across three datasets. Across the three datasets—tgbn-trade-TTE, tgbn-genre-TTE, and tgbn-reddit-TTE. We observe that \method~(w/ Self-att) generally achieves lower MAE compared to \method~(w/ MLP).
The results indicate that the self-attention mechanism in \method's Graphormer-like structure is more effective for evaluating dynamic graph models compared to a simple MLP backbone. This is because Graphormer-like structure can capture the structure information without edges.

\subsection{In-depth Analysis of \method.}
\subsubsection{Impact of Number of Simulated Dynamic Graphs.}
\begin{figure}[h]
    \centering
    \includegraphics[width=0.9\linewidth]{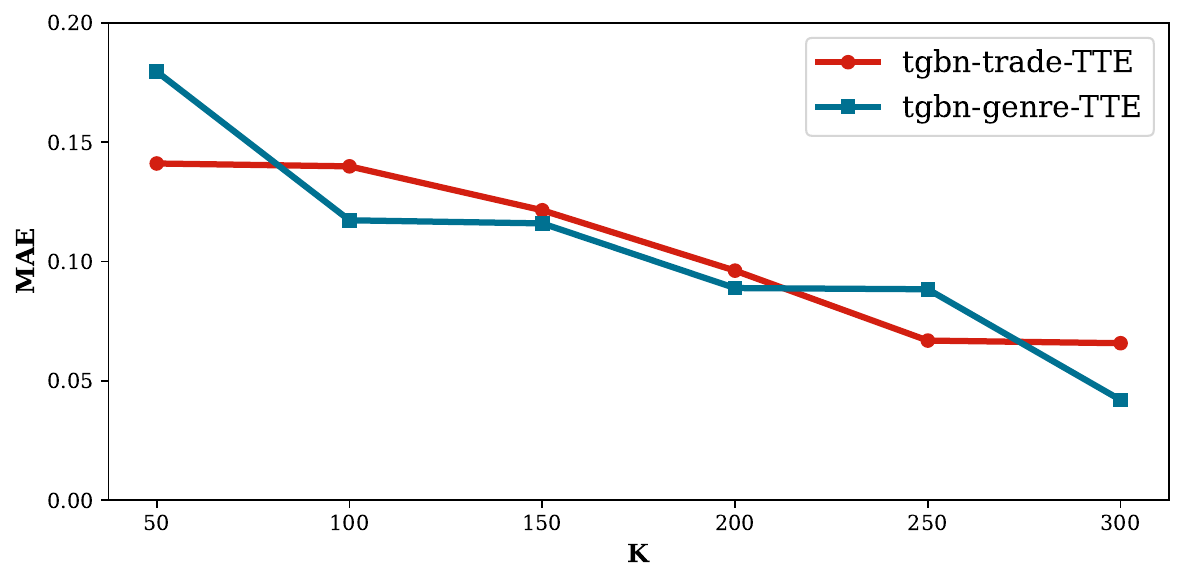}
    \caption{MAE of \method~with different number of simulated graphs trained by TGN.}
    \label{fig:simulated-graph-number}
\end{figure}
We evaluate the effects of different numbers of simulated dynamic graphs \((K)\) on \method. Specifically, we use the number of simulated graphs for \(K=[50,100,150,200,250]\) to train the proposed \method~for evaluating the TGN that is well-trained on tgbn-trade and tgbn-genre, making inferences on unseen tgbn-trade-TTE and tgbn-genre-TTE. We consider the MAE across eight dynamic test graphs for each dataset. Generally, we observe that the MAE decreases as the number of simulated graphs increases, indicating that the proposed \method~can effectively learn to make predictions by training with more simulated graphs.

\subsubsection{Running Time Comparison.}
We evaluate the time-consuming between proposed \method~and GNNEvaluator. Table~\ref{tab:time} presents a comparison of the time consumed per epoch by GNNEvaluator and our proposed \method~across two datasets: tgbn-trade-TTE and tgbn-genre-TTE. We observe that the running time of \method~is slightly higher than that of GNNEvaluator; however, \method~consistently demonstrates better evaluation performance, as shown in earlier sections of the analysis.
However, we must also consider that GNNEvaluator uses only the edges from the last timestamp. This implies that if we want to incorporate more timestamps, the complexity of this method will also increase. Therefore, \method~remains a strong choice for the DGNN model evaluation task. Note that this experiment was conducted on an Nvidia A100 GPU with 40 GB of memory.

\section{Conclusion and Future Work}
In this work, we tackled a new challenge: test-time DGNN model evaluation for understanding and assessing the performance of well-trained DGNNs on unseen dynamic graphs. We proposed a two-stage approach. First, we generate a simulated dynamic graph set capture discrepancies across different dynamic graph distributions. Using this, \method~is trained to predict node property performance, such as NDCG, for well-trained DGNN models on unseen dynamic graphs. Extensive experiments with real-world unseen dynamic graphs demonstrate the effectiveness of our proposed method for DGNN model evaluation. Our approach assumes that the output space remains consistent across training and testing graphs, despite the presence of covariate shifts. In future work, we aim to relax this assumption and address a wider range of natural real-world dynamic graph data distribution shifts.

\section*{Acknowledgments}
    This research is supported by the U.S. National Science Foundation and Australia CSIRO joint project: NSF-CSIRO: Towards Interpretable and Responsible Graph Modeling for Dynamic Systems (IIS-2302786). This work has also been supported by the Australian Research Council (ARC) under grants FT210100097 and DP240101547.
\bibliographystyle{IEEEtran}
\bibliography{IEEEexample}
\end{document}